\documentclass{article} % For LaTeX2e

\usepackage[nonatbib,final]{nips_2016}

\usepackage[utf8]{inputenc} % allow utf-8 input
\usepackage[T1]{fontenc}    % use 8-bit T1 fonts
\usepackage{hyperref}       % hyperlinks
\usepackage{url}            % simple URL typesetting
\usepackage{booktabs}       % professional-quality tables
\usepackage{amsfonts}       % blackboard math symbols
\usepackage{nicefrac}       % compact symbols for 1/2, etc.
\usepackage{microtype}      % microtypography
\usepackage{tabu}

\usepackage{graphicx}
\usepackage{subcaption}

\title{LSUN: Construction of a Large-Scale Image Dataset using Deep
  Learning with Humans in the Loop}

\author{
  \begin{tabu} to 0.8\textwidth {*4{X[c]}@{}}
    Fisher Yu & Ari Seff &  Yinda Zhang &
    Shuran Song
    \\
    \\
    \multicolumn{2}{c}{Thomas Funkhouser} &
    \multicolumn{2}{c}{Jianxiong Xiao}
        \\
        \\
        \multicolumn{4}{c}{\textnormal{Princeton University}}
  \end{tabu}
  }

\begin{document}

\maketitle

\setlength{\textfloatsep}{0.75\textfloatsep}
\setlength{\intextsep}{0.75\intextsep}
\setlength{\floatsep}{0.75\floatsep}

\begin{abstract}
While there has been remarkable progress in the performance of visual
recognition algorithms, the state-of-the-art models tend to be
exceptionally data-hungry. Large labeled training datasets, expensive
and tedious to produce, are required to optimize millions of
parameters in deep network models. Lagging behind the growth in model
capacity, the available datasets are quickly becoming outdated in
terms of size and density.  To circumvent this bottleneck, we propose
to amplify human effort through a partially automated labeling scheme,
leveraging deep learning with humans in the loop.  Starting from a
large set of candidate images for each category, we iteratively sample
a subset, ask people to label them, classify the others with a trained
model, split the set into positives, negatives, and unlabeled based on
the classification confidence, and then iterate with the unlabeled
set.  To assess the effectiveness of this cascading procedure
and enable further progress in visual recognition research, we
construct a new image dataset, LSUN.  It contains around one million
labeled images for each of 10 scene categories and 20 object categories.  We
experiment with training popular convolutional networks and find that
they achieve substantial performance gains when trained on this dataset.

\end{abstract}

\vspace{-0.05in}
\section{Introduction}
\vspace*{-0.075in}
%In the past three years,
High capacity supervised learning algorithms, such as deep
convolutional neural networks \cite{lecun1998gradient,AlexNet},
%have enabled significant progress 
have led to a discontinuity in visual recognition over the past four
years and continue to push the state-of-the-art performance (e.g. \cite{MSRA,wu2015deep,GoogleNorm}). These models
usually have millions of parameters, resulting in two consequences.
On the one hand, the many degrees of freedom of the models allow for extremely impressive description power; they can learn to represent complex functions and transformations automatically from the
data. On the other hand, to search for the optimal settings for a
large number of parameters, these data-hungry algorithms require
a massive amount of training data with humainitiallced labels
\cite{deng2009imagenet,fei2007learning,xiao2010sun,zhou2014learning}.

Although there has been remarkable progress in improving deep
learning algorithms (e.g. \cite{MSRA,GoogLeNet}) and developing high
performance training systems (e.g. \cite{wu2015deep}), advancements
are lacking in dataset construction.  The ImageNet dataset
\cite{deng2009imagenet}, used by most of these algorithms, is
7 years old and has been heavily over-fitted \cite{BaiduScandal}.  The 
recently released Places \cite{zhou2014learning} dataset is not much larger.
The number of parameters in
many deep models now exceeds the number of images in these datasets.
%stretching the performance under the risk of the lack of bounded
%constraints \cite{Intriguing,Fooled,AdversarialExamples}.  % Clarify...
While the models are getting deeper
(e.g. \cite{VGGnet,GoogLeNet}) and the accessible computation power is
increasing, the size of the datasets for training and evaluation is
not increasing by much, but rather lagging behind and hindering further
progress in large-scale visual recognition.

Moreover, the density of examples in current datasets
is quite low.  Although they have several million images,
they are spread across many categories, and so there
are not very many images in each category (e.g., 1000 in ImageNet). As a result,
deep networks trained on them often learn features that are noisy
and/or unstable~\cite{Intriguing,Fooled}. To address this problem,
researchers have used techniques that augment the training sets with
perturbations of the original images~\cite{wu2015deep}.  Those methods
add to the stability and generalizability to trained models, but without
increasing the actual density of novel examples.  We propose an
alternative: an image dataset with high density. %but few categories. - why would we advertise few categories?
We aim for a dataset with $\sim10^6$ images per category, which
is around 10 times denser than
PLACES~\cite{zhou2014learning} and 100 times denser than
ImageNet~\cite{deng2009imagenet}.

Although the amount of available image data on the Internet is
increasing constantly, it is nontrivial to build a supervised dataset that
dense because of the high costs of manual labeling.  Clever methods
have been proposed to improve the efficiency of human-in-the-loop
annotation (e.g., \cite{branson2010visual,deng2014scalable,russakovsky2015best,vijayanarasimhan2008multi,wah2011multclass}).
However, manual effort is still the bottleneck -- the
construction of ImageNet and Places both required more than a year of
human effort via Amazon Mechanical Turk (AMT), the largest
crowd-sourcing platform available on the Internet. If we desire a
$N$-times bigger/denser dataset, it will require more than $N$ years of
human annotation. This will not scale quickly enough to support
advancements in deep visual learning. Clearly, we must
introduce a certain amount of automation into this process in order to
let data annotation maintain pace with the growth of deep models.

%However,
%we need to be extremely cautious in ensuring the accuracy of labels
%obtained (semi-)automatically. In addition, humans make mistakes
%quite often, especially when performing repetitive and tedious jobs
%such as data labeling.
% figure to point out errors in places annotation.
% Thus, a systematic way to check human labeling accuracy is a
% necessity. %hopefully even to surpass human accuracy by combining
%automatic algorithms and human annotation. - Not sure %about this

In this paper, we introduce an integrated framework using deep
learning with humans in the loop
%combining deep learning and AMT crowd sourcing 
to annotate a large-scale image dataset. The key new idea is a labeling propagation
system that automatically amplifies manual human effort.  We study strategies for
selecting images for people to label, interfaces for acquiring labels quickly,
procedures for verifying the labels acquired, and methods for amplifying the labels by
propagating them to other images in the dataset.  We have used the system to
construct a large scale image database, ``LSUN'', with 10 million labeled images
in 10 scene categories and 59 million labeled images in 20 object categories.\footnote{We are continuing to collect and
annotate data through our platform at a rate of 5 million images and 2 categories
per week}.

One of the main challenges for a semi-automatic approach
is achieving high precision in the final labeled dataset.  
Our procedure uses statistical tests to ensure labeling quality,
providing more than 90\% precision on average according to
verification tests.  This is slightly lower than manual annotation,
but still good enough to train
high performance classification models.  During experiments with
popular deep learning models, we observe
substantial classifier performance gains when using our larger training
set.  Although the LSUN training labels may contain some noise, it
seems that training on a larger, noisy dataset 
produces better models than training on smaller, noise-free datasets.

% Training vs. test set precision?

%objects too
%    LSUN for scene classification with more than 10 millions images.

\vspace{-0.05in}
\section{Overview}
\vspace*{-0.075in}
An overview of our labeling pipeline is shown in Figure~\ref{fig:pipeline}.
For each category, we start by collecting a large set of candidate images
using keyword search (usually $\sim10^7-10^8$ images).   Then,
we iterate between asking people to label a small subset, training a
classifier on that subset, asking the classifier to predict labels
and confidences, and then selecting a small candidate set for further
consideration.

In many ways, this process is similar to active learning
\cite{settles2010active,tong2002support,collins2008towards,vijayanarasimhan2008multi}.  However,
there is a key difference: we aim to get correct labels for all
examples in a specific set of images rather than learning a
generalizable classifier.  The main implication of this difference is
that it is acceptable for our system to over-fit a model to adapt
locally to its target set, without considering generalizability outside
the set.

Our iterative method learns a cascade of over-fit classifiers,
each of which splits a set of images into positive, negative, and
unlabeled subsets.  The positive examples are added to our dataset;
the negative examples are discarded; and the unlabeled subset is
passed to the next iteration for further processing.  The cascade
terminates when the unlabeled subset is small enough that all images
can be labeled manually.  If the unlabeled subset shrinks by a
significant factor at each iteration, the process can label a very
large data set with a practical amount of human effort.

The following sections describe the main components of our system,
with detailed descriptions of how we collect data, learn a deep
network classifier for each iteration, design an interface for manual
labeling, and control for label quality.  These descriptions
are followed by results of experiments designed to test whether the
collected dataset is helpful for image classification.

\vspace{-0.05in}
\section{Data Collection}
\vspace*{-0.075in}
The first step in constructing a large-scale image database is collecting the 
images themselves. As in previous work, we leverage existing image search
engines to gather image URLs. Here, we use Google Images 
search. By querying with appropriate adjectives and 
synonyms, we can bypass the usual limit on search results and obtain nearly 
100 million relevant URLs for each category.

Similar to SUN and MS COCO, our target database will contain images from both scene and object categories. To generate queries for scene categories, 696 common adjectives relevant to scenes (messy, spare, sunny, desolate,
etc.), manually selected from a list of popular adjectives in English
(obtained by \cite{zhou2014learning}), are combined with each scene
category name. Adding adjectives to the queries
allows us to both download a larger number of images than what is available
in previous datasets and increase the diversity of visual appearances. To further increase the number of search results per category, we set the time span for each query to three days and query all
three-day time spans since 2009. We find that a shorter time span
seldom returns more useful results. We then remove duplicate URLs and
download all accessible images (some URLs may be invalid or the images not properly encoded). These same methods are used to collect image URLs for object categories, except the queries use a different set of object-relevant adjectives. As an example of our querying results, we obtain more than 111 million unique URLs for images relevant to
``dog''.

To date, more than 1 billion images have been downloaded. An initial
quality check is performed before labeling these images. Only
images with smaller dimension greater than 256 are kept. After the
check, around 60\% of the images are kept. Unlike previous
systems ~\cite{xiao2010sun,zhou2014learning}, we don't remove
duplicates (in terms of image content)
in this initial pool of images as it is very expensive to remove
duplicates while avoiding overkill on tens of millions of non-duplicate images. In
addition, our pipeline requires only a small subset of this pool (with rare duplicates) to be manually labeled. Therefore leaving duplicates in the original image pool does not increase the cost of manual labeling. The
duplicate images usually exhibit various compression qualities,
sizes, and even croppings. They augment our dataset naturally, and it
will be up to algorithm designers how to utilize the
duplication. All images from this initial pool will be released for
unsupervised image analysis.

%In the following two sections, we first describe our labeling procedure (leveraging deep learning with humans in the loop), and then we detail how we massively crowd-source human labels from AMT.

\vspace{-0.05in}
\section{Deep Learning with Humans in the Loop}
\vspace*{-0.075in}
\begin{figure}[t]
\centering
\includegraphics[width=1\linewidth]{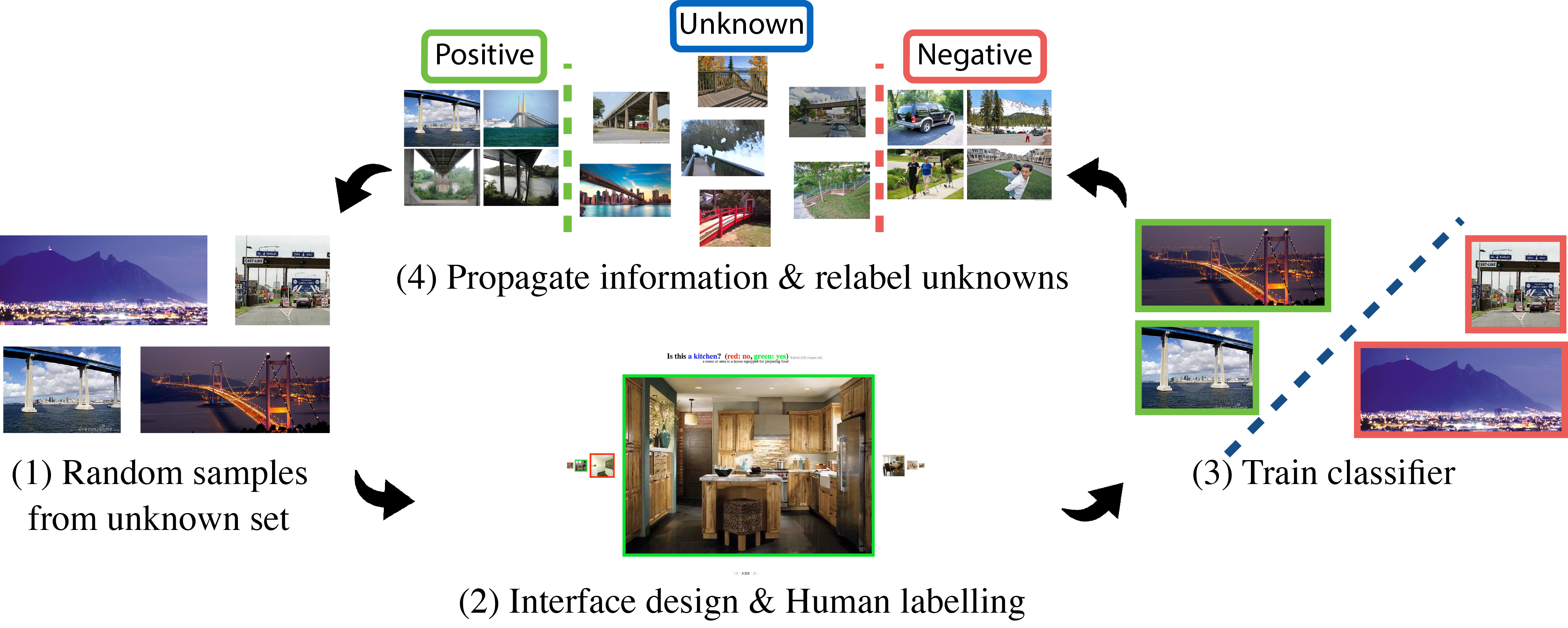}

\vspace{-3mm}

\caption{An overview of our pipeline. To annotate a large number of 
images, we first randomly sample a small subset (1) and
  label them through crowdsourcing using our Amazon Mechanical Turk 
interface(2). This small labeled subset is then utilized to
  train a binary classifier with deep learning feature (3). Then, we
  run the binary classifier on the unlabeled images. The images
  with high or low scores are labeled as positive or
  negative automatically, while the images ambiguous to the classifier
  are fed into the next iteration as the unlabeled images (4). The
  pipeline runs iteratively until the number of images remaining in the 
unknown set
  is amenable to exhaustive labeling.}

\vspace{-3mm}

\iffalse
\caption{The overview of our pipeline. (a) We query and download
hundreds of millions of images from the Internet. We also do some
checking to remove the bad images. (b) A testing set is uniformly
sampled from the download images and labeled by AMT. This set has
multiple purposes as described in Section~\ref{sec:test_set}. (c) We
first use clustering to find clusters of good and bad images. (d) Then
various SVM models are trained by iterative sampled image set to split
positive and negative images. (e) Finally, we use fine-tuning on
AlexNet to train proper feature representation for the hard to
separate cases and split the remaining sets further.}
\fi

\label{fig:pipeline}
\end{figure}

Although we can collect around 100 million relevant images for each
category, 
the time and human effort required for manual labeling is prohibitive. We 
observe that existing methods for image classification tasks obtain 
impressive results, as shown in the
recent ImageNet challenges~\cite{russakovsky2015imagenet}. We 
therefore propose to use image classification algorithms in combination 
with humans to label the images semi-automatically, amplifying human
effort. 

We treat the labeling process for each category
as a binary classification problem -- i.e., every image
downloaded for the category is labeled as either 
positive or negative.

%While we can adapt convolutional network features or fine-tune 
%previously trained models to this domain, it is not initially clear which 
%models or number of training images will optimize the performance of 
%our pipeline. To address this problem, we develop an iterative method 
%with humans
%kept in the loop to provide additional labeled training images for each
%iteration.

Starting with the large pool of unlabeled 
images. We randomly sample a small subset of images from this pool 
and acquire labels for them from AMT workers (see section \ref{crowd} for details). 
This labeled set is then randomly split into a 
training and testing set. We train a classifier on the training set and then 
run it on both the labeled testing set and the full unlabeled pool. By 
examining the scores output by the classifier on the labeled testing 
set, we can determine score ranges which delineate either confident 
scores 
corresponding to easy images or weaker scores corresponding to more 
difficult or ambiguous images. 

To define these score ranges, we compute two thresholds: the score 
above which 95\% of the images are ground truth positives and the score 
below which there are only 1\% of the total positive images. Both in the 
labeled subset and the unlabeled pool, images scoring above or 
below this threshold are labeled as positive or negative, respectively. All 
images scoring in between these two thresholds, i.e., the more 
challenging images, are sent to the next iteration. Thus, during each 
iteration, we label a portion of images with confident scores as positive 
and also remove highly probable negative images while
maintaining a good recall for the final labeled set. 

This process is repeated identically for each iteration except that, 
beginning with the second iteration, the randomly sampled subset for 
manual labeling will be combined with the ambiguous portion of the 
previous iteration's labeled subset. Therefore, because later iterations 
have a 
larger number of difficult training examples, the later
classifiers
can outperform previous classifiers, confidently assigning labels to some 
of
the images in the unlabeled pool. Then, the images still lacking 
confident labels are passed to the next
iteration. This process is illustrated in Figure~\ref{fig:pipeline}.
The next two sections will discuss details regarding performing 
classification and sampling images in each iteration.

%\vspace*{-2mm}
%\subsection{Classification}
%\vspace*{-0.05in}

\vspace*{1mm}
\noindent{\bf Classification:}
We encounter a tradeoff between computation time and 
performance when selecting classification models for our pipeline. Although the most accurate classification model will limit the number of iterations necessary to label the entire image pool, we may
not be able to afford its running time on tens of millions of
images. Also, the current successful classification models are all
based on convolutional networks. Better models are usually deeper and of higher capacity, requiring more training images.

To make the tradeoff, we assess several different classification models. We find that, for the initial image set, the image
representation from the last fully connected layer in AlexNet can
be used to separate the images effectively. By using a multi-layer
perceptron (MLP) with two hidden layers, we can remove more than half 
of the
images from the initial set. However, after two iterations with a MLP,
this feature space is no longer effective. So we then adapt
pre-trained GoogLeNet~\cite{GoogLeNet} by fine-tuning, which can
continuously benefit from the additional training images sampled at
each
iteration. The learning rate of fine-tuning starts with 0.001 and
drops to 0.0001 after 40,000 iterations. The total number of
iteration is 60,000. The mini batch size is 80. In the testing phase
during early iterations, we
only pass the central crop of each image to the model. After 10
labeling iterations, we train the network for the labeling iterations
divided by 10 more training iterations. Also, we pass
10 crops to the network and average the score output. Although
evaluating with 10 crops is 10 times more expensive, far fewer images remain after 10 labeling iterations.

To select the best models from the training process, we evaluate the
models on a validation set and compare the models by accuracy. Because the validation set has a low ratio of positive to negative images we must carefully select the accuracy metric. Given a set of images for validation, we measure two scores of
accuracy by assigning different weights to the images. For one score, all
the images have the same weight. In the second score, the sums of the
weights of positive and negative images are equal, with the set of
positive images and the set of negative images each being weighted
uniformly. We select the model that removes more images in the testing set.

%\vspace*{-2mm}
%\subsection{Image Sampling}
%\vspace*{-0.05in}

\vspace*{2mm}
\noindent{\bf Image Sampling:}
To improve the classifier at each iteration, we sample 40 thousand
images uniformly randomly from the unlabeled pool. 10 thousand are reserved for testing the
classifier confidence. 5 thousand are used for validation to pick the
best models from the training procedure. The remaining images are used for
training. At the end of each iteration, after the classifier labels the
positive images and removes negative images, the labels of all the
images in the remaining set are kept for training in the next
iteration. We have also experimented with sampling 80 or 120
thousand images per
iteration but it did not improve the classifier results significantly
during the initial stages.

\vspace{-0.05in}
\section{Crowd Sourcing} \label{crowd}
\vspace*{-0.075in}
%\begin{figure}[t]
%\includegraphics[width=\linewidth]{img/interface.pdf}
%\vspace{-4mm}
%\caption{Our AMT Labeling Interface.}
%\label{fig:mturk_interface}
%\end{figure}

\begin{figure}[t]
\vspace{-3mm}
\centering
\includegraphics[width=\linewidth]{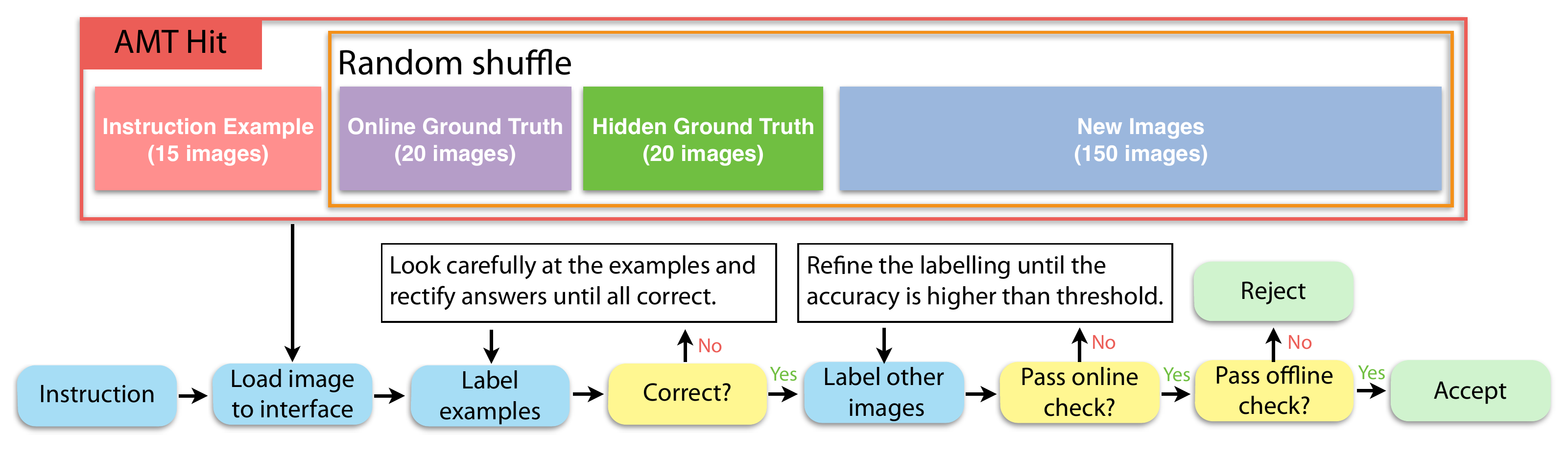}
\vspace{-3mm}
\caption{Life time of a HIT on AMT.}
\label{fig:mturk_pipeline}
\vspace{-5mm}
\end{figure}

A critical part of our pipeline is obtaining high quality manual 
annotations (from humans in the loop).
Of necessity is the ability to obtain many labels, quickly and at minimal cost.
To this end, we use Amazon Mechanical Turk (AMT), the largest 
crowdsourcing platform available.
However, annotations
obtained from AMT can vary in quality.
Here we present a series of mechanisms and measures to
ensure high quality annotations.

%\vspace{0mm}
%\subsection{Interface Design}
%\vspace*{-0.05in}

\vspace*{2mm}
\noindent{\bf Interface Design:}
Built on the labeling interface from \cite{zhou2014learning},
our interface allows human annotators to go through
the images one by one.
As shown in the supplemental material,
a single image is displayed on the screen at a time along with a question 
asking if the image content fits a particular category (e.g., ``Is this a 
kitchen'').
A definition of the category is also provided. Small thumbnails are shown
for the previous three images to the left and upcoming three images to the
right.
By pressing the arrow keys, a worker can navigate through the images, with 
the default answer of each image set to ``No''.
The worker can press the space bar to toggle the current answer (encoded by 
the color of the
boundary box).

We observe that scene images % an object images?
can be complicated, and a larger view of the images is often helpful for 
comprehension. Likewise we display one image at a time, maximizing the 
image size to fill the window. To make sure the
window can fully utilize the worker's screen, we provide a simple
button that forces the labeling window to enter fullscreen mode.

%\vspace{-2mm}
%\subsection{Instructions}
%\vspace*{-0.05in}

\vspace*{2mm}
\noindent{\bf Labeling Instructions:}
There is some ambiguity inherent to the task of assigning scene and object 
categories to images. For example, how much of a person's body must be 
visible for an image to be considered positive for the category 
``person''? If only a face is shown, does this count? What if only an arm is 
shown? This ambiguity can be amplified by the fact that AMT workers come 
from a variety of countries and cultural backgrounds; their preconceived 
notions of particular categories may differ from ours. In order to thoroughly clarify the 
range of image content we desire for each scene and object, we include an 
instructions page displaying example images, their ground truth labels, and 
explanations for the labels.

To gather effective examples for each category, we first address obvious 
corner cases. For example, for the ``car'' category, we would like trucks and 
buses to be excluded as these will later form separate categories. Therefore, 
we specify that trucks and buses should be marked as negative on the ``car'' 
instruction page. In addition, for each object category, we decide which parts 
of the object are significant enough that when visible, they make the object's 
identity clear. For example, when only one wheel is visible in an image, there 
is not enough information to confidently assert the presence of a bicycle. 
While we acknowledge that these decisions are ultimately subjective, by 
keeping these instructions consistent for all the workers we can build a 
dataset that cleanly abides by these class definitions.

After addressing corner cases, we run some test hits on AMT to discover any 
other possible sources of confusion. For each category, we sample
30,000 images from all the images and label each image by 3 people
on the AMT and check the images with conflicting labels.  We manually
select a subset of them to seed a pool of
example corner cases, from which we randomly sample 15
to show the AMT worker before each hit.

%We found AMT workers often label goats as sheep.
%In consulting other datasets such as PASCAL 
%VOC, we found that goats were heavily prevalent among the ``sheep'' 
%category images. We decided to allow goats in this category as well. 
%Likewise, the sheep instructions page was updated to include examples of 
%goat images as being positive cases.

In addition to category-specific examples, we also provide the workers with 
examples of general types of images we do not want to include in our 
dataset. For example, as we wish to build a natural image dataset, we advise 
the workers to mark as negative any computer-generated or cartoon imagery. 
Images with obtrusive text overlay (where solid font text occludes the object 
of interest) and photo collages are similarly marked as negative. If an object 
of interest is printed on a magazine/book/TV screen within an image, we 
consider this to be negative. In these cases it is actually the magazine/book/
TV screen that is physically present within the scene, not the object of 
interest.
% explain why we allow text but not other unnatural entities like collages

%% \textbf{List of object category examples} (to be moved to supplementary 
%% material):

%\vspace{-2mm}
%\subsection{Quality Control}
%\vspace*{-0.05in}

\vspace*{2mm}
\noindent{\bf Quality Control:}
In any crowdsourcing platform like AMT, the quality of work can vary from 
worker to worker.
We take the approach of redundant labeling and enforced instruction. For 
each
image, we gather labels from two workers and only keep doubly confirmed
labels.

%% 1) Clear Instruction
%% 2) Tutorial mode
%% 3) Full screen for easy recognition

Given a set of images that need to be labeled, we divide them into
groups of 150 images.
In each AMT HIT, we have the human annotator label 205 images.
150 of these are the actual images we are interested in and the remaining
55 are included for quality control purposes. The workers normally finish the 
hits
within 5 minutes.
Having tested a variety of hit lengths, we found this to be a good trade-off 
between efficiency and quality control.
%to be a proper workload for a HIT.

%Our workers come from 66 countries, with 

%% Although we don't have to label every image with our pipeline, we
%% still need large amount of labels to train our models. We resort to
%% AMT to crowd source image labels.

Figure \ref{fig:mturk_pipeline} shows the steps for the life time of a
HIT.
First, 
we show an instruction page to explain the task and questions with
positive and negative examples. 
Although it is useful and necessary,
we find that some workers may not actually read through the
instruction very carefully, especially the detailed definition and the
positive and negative examples.
They may also forget about the subtlety for some categories.
Therefore, we set the first 15 images in each HIT to test whether the
workers understand the instruction. The images are from a set of
examples representing typical category images and common mistakes. If a
worker gives a wrong label to one of the example image, a pop window
will show up and block the labeling interface until the worker fixes
the label to that image. 
After the 15 tutorial images,
they will go on to label without immediate feedback per image.

%% Quality control

%% 1) online ground truth check
%% 2) hidden ground truth check
%% 3) Iterative labeling testing set to find good set of ground
%% truth. The importance of each ground truth image will be collected
%% by
%% workers' real performance and dynamically balanced sample rate.

%To improve the quality of the submitted hits, we have 
We embed 40 images sampled from hundreds with
known ground truth in each HIT to measure the labeling quality. 
20 of them are online, meaning that their quality is checked before
the HIT
result is submitted to the server. If the worker can't pass the online
check, they are not allowed to submit the labeling results and advised
to revise their labeling results until they can pass the online
check. 
The other 20 of those images are
checked after the HIT is submitted. They are used to check whether the
worker hack our Javascript code to pass the online check to make sure
they exhaustively look for online
ground truth to pass the submission test. The labels of these images
are not sent to the worker's browser, so they can't hack our interface
to submit bad results. The labeling accuracy threshold for ground
truth is 90\% for the online and 85\% for the hidden.

%We maintain hundreds of ground truth images and dynamically sample
%them for each HIT.

%Together with the examples, the ground truth check is to make sure the
%workers submit high quality labels.  
%We also use them to prevent the
%common mistakes.  We try to understand the kinds of common mistakes
%from the wrong labels in the testing set. When we label the images in
%the training set, we continue adding more images to examples and
%ground truth when necessary.

\vspace{-0.05in}
\section{Results}
\vspace*{-0.075in}
We are running this labeling pipeline continuously and have now
collected enough data to evaluate how well it is performing and
study its potential impact on visual recognition.

%We have constructed an initial version of LSUN with
%millions of labeled images in each scene category.
%We experiment with popular convolutional networks using our dataset, and obtain a
%significant performance gain with the same model but trained using our
%bigger training set.

% Given the our pipeline, we want to know the quantity and quality of
% the
% resulting dataset. In addition, we investigate how effectively our
% pipeline is and how the current computer vision models work on our
% dataset.

%%%%%%%%%%%%%%%%%%%%%%%%%%%%%%%%%%%%%%%%

%\vspace{-2mm}
%\subsection{Dataset Statistics}
%\vspace*{-0.05in}

\begin{figure}[t]
  \centering
    \includegraphics[width=\textwidth]{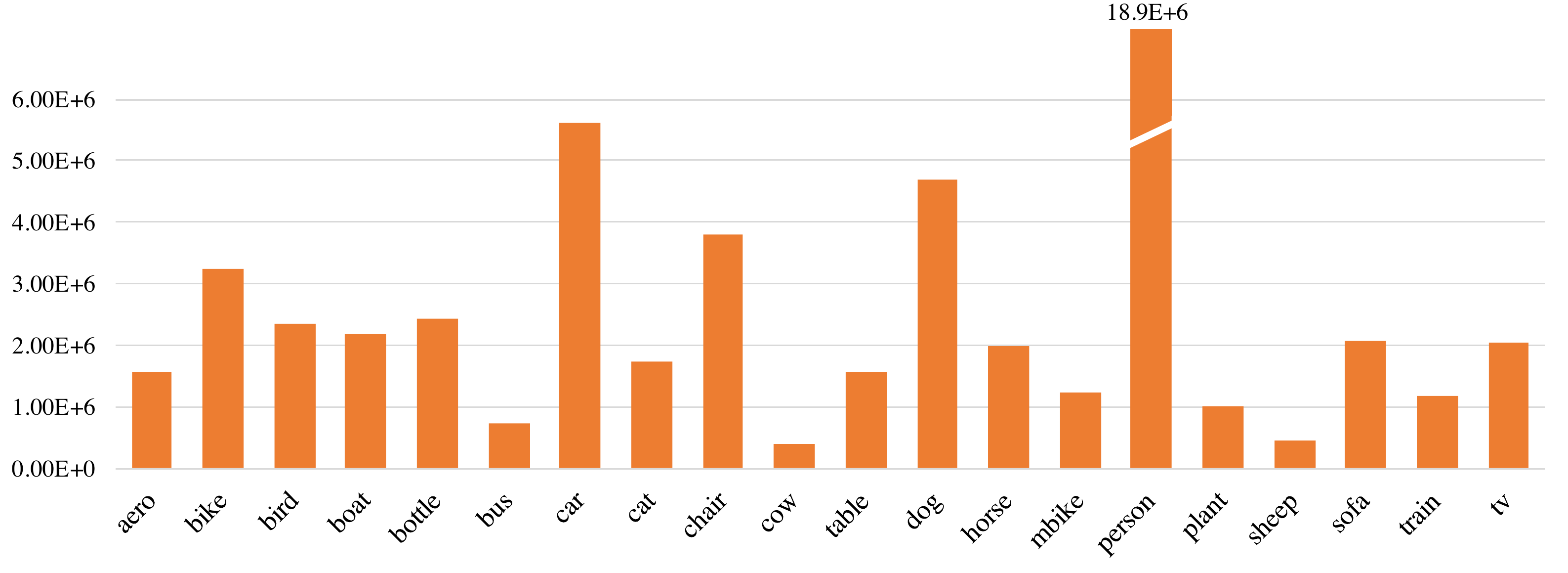}
   \caption{Number of images in our object categories. Compared to
    ImageNet dataset, we have more images in each category, even
    comparing only basic level categories.}
\label{fig:numbers}
        \end{figure}

\vspace*{2mm}
\noindent{\bf Dataset Statistics:}
So far, we have collected 59 million images for 20 object categories
selected from PASCAL VOC 2012 and 10 million images for 10 scene
categories selected from the SUN database.
Figure~\ref{fig:numbers} shows the
statistics of object categories. The number of images used in ImageNet
classification challenge is about 1.4 million, which is used to train
recent convolutional networks. Most of our object
categories have more images than the whole challenge. Even if we only
consider the basic level categories in ImageNet such as putting all
the dogs together, our dataset is still much denser.

%% For most of the categories, of which we have far
%% fewer, our dataset is more than 10 times denser (e.g., XXX vs. XXX for
%% bedroom). Similarly, our object categories are approximately 1000 times
%% fewer and denser than ImageNet (e.g., density is XXX vs. XXX for chair).

%%%%%%%%%%%%%%%%%%%%%%%%%%%%%%%%%%%%%%%%

%\vspace{-2mm}
%\subsubsection{Label Precision}
%\vspace*{-0.05in}

\begin{figure}[hbtp]
  \centering
  \begin{subfigure}[b]{0.48\textwidth}
    \includegraphics[width=\textwidth]{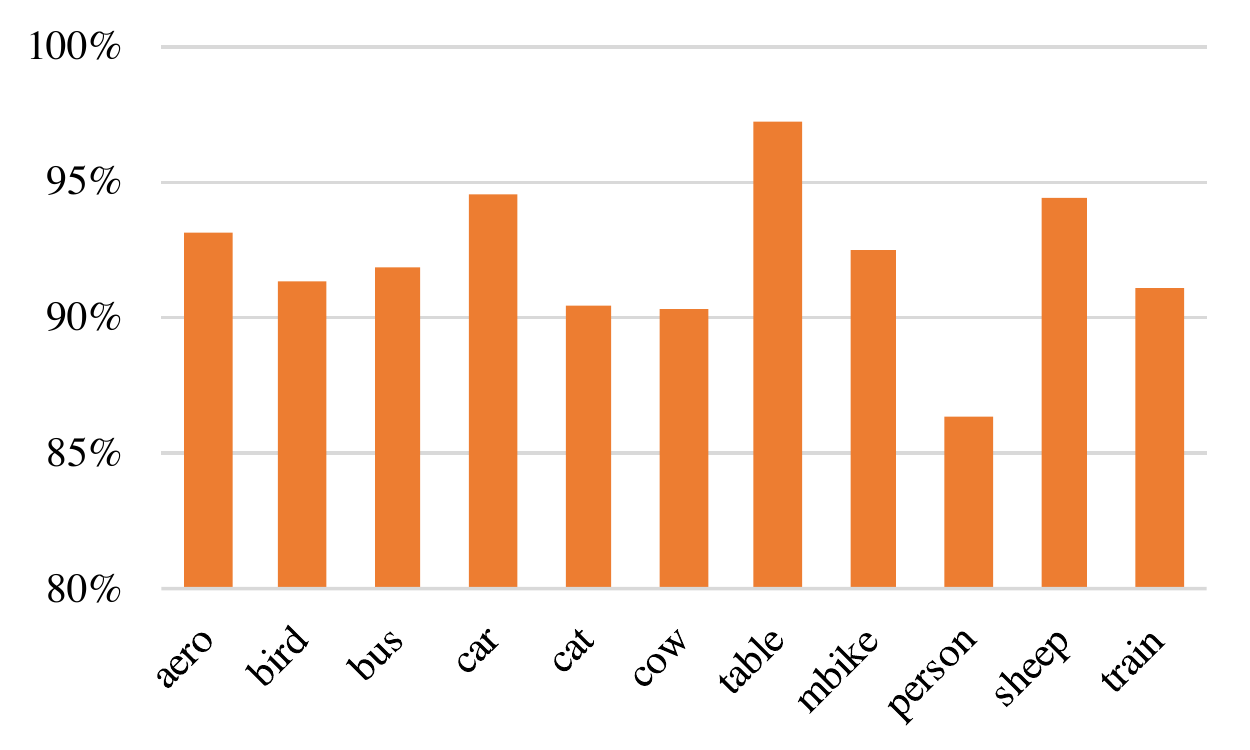}    
    \caption{Precision of our labels}
    \label{fig:pr_precision}
  \end{subfigure}
    \begin{subfigure}[b]{0.48\textwidth}
    \includegraphics[width=\textwidth]{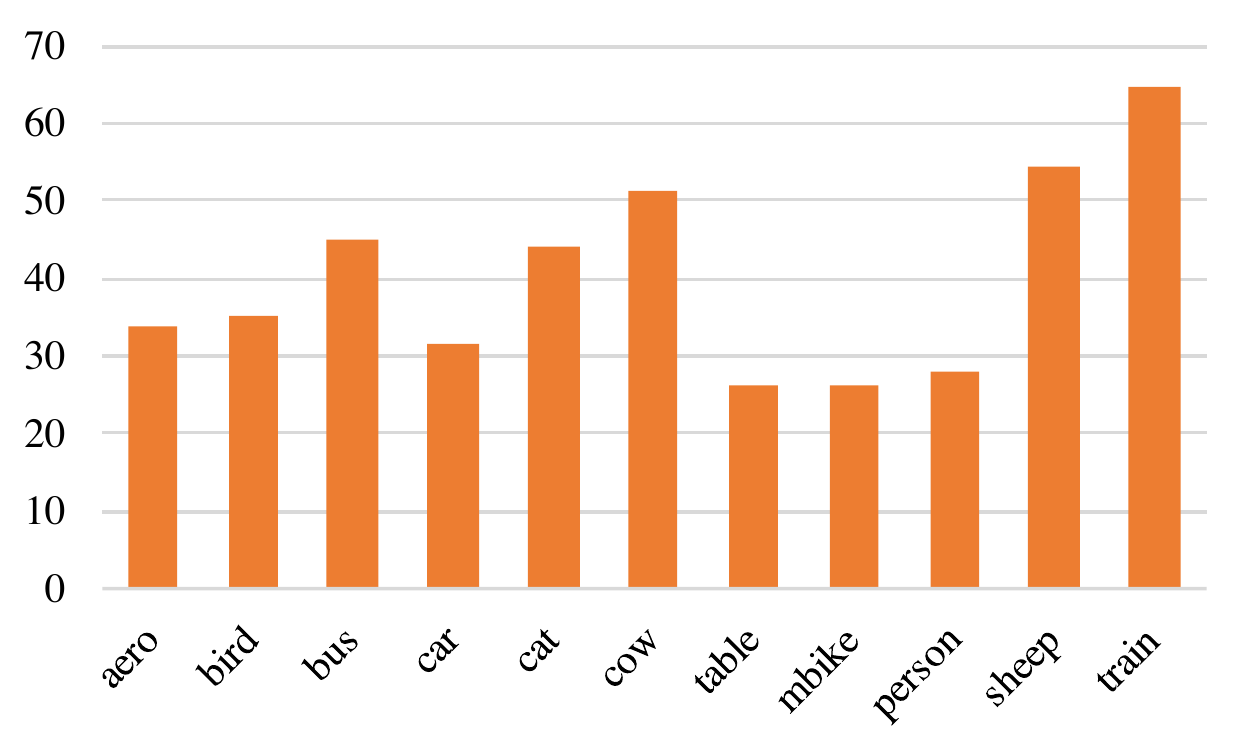}
    \caption{Amplification of human labels}
    \label{fig:pr_reduction}
  \end{subfigure}
\caption{Labeling precision and reduced labeling efforts of our
  pipeline. (a) shows the precision of labels from 11 object
  categories. The precision is normally around 90\%. (b) shows how
  much labeling effort is saved with our
  pipeline.  On average the ratio between all images 
  and the number of human-provided labels is around 40:1.}
\label{fig:precision_reduction}
\end{figure}

\vspace*{2mm}
\noindent{\bf Label Precision:}
We have run a series of experiments 
to test the labeling precision of this dataset.
We sampled 2000 images from 
11 object categories and label them with 
completely manual pipeline using trained experts (AMT labeling
is not reliable enough for this test).  The final
precision of those categories is shown in
Figure~\ref{fig:pr_precision}a. We observe that normally, the precision
is around 90\%, but it varies across different categories. The major
mistakes in our labels are caused by toys, computer rendered images,
photo collages, and human edited photos.  The inclusion of
computer-rendered images is the most serious for the person
category. Although we have tests for classifier confidence, there are
lots of confusing cases for human labelers on these images,
which subsequently confuse the statistical tests.

%%%%%%%%%%%%%%%%%%%%%%%%%%%%%%%%%%%%%%%%

%\vspace{-2mm}
%\subsubsection{Effort Amplification}
%\vspace*{-0.05in}

\vspace*{2mm}
\noindent{\bf Effort Amplification:}
We next checked how much our pipeline
amplifies human labeling effort -- i.e., the ratio between the number
of images put into a positive or negative set versus the number of images
labeled by people. The result is
shown in Figure~\ref{fig:pr_reduction}. It shows that our pipeline
amplifies human efforts by 40 times on average. In the case of the train
category, people labeled less than 1/60th of the images.

%%%%%%%%%%%%%%%%%%%%%%%%%%%%%%%%%%%%%%%%

%\vspace{-2mm}
%\subsection{Model Performance}
%\vspace*{-0.05in}

\begin{figure}[hbtp]
\centering
\includegraphics[width=\textwidth]{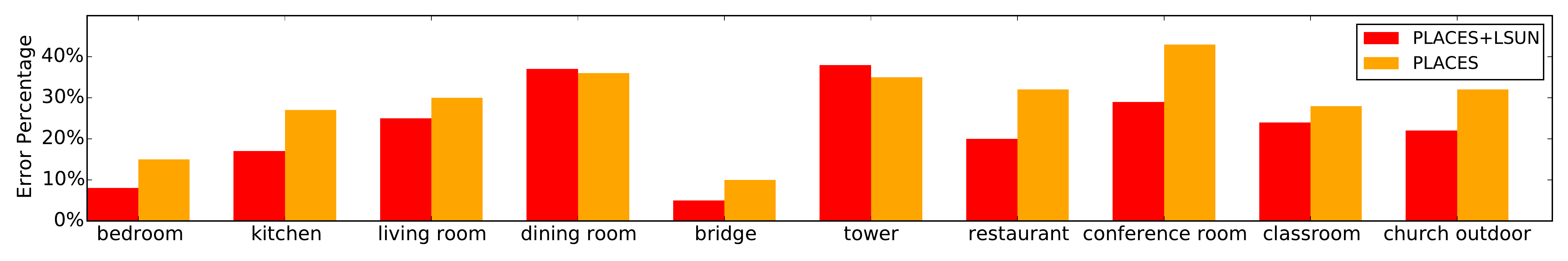}
\caption{Comparison of classification error by training AlexNet with PLACES
vs. LSUN and PLACES.  Lower errors are better.}
\label{fig:comp_places_error}
\end{figure}

\vspace*{2mm}
\noindent{\bf Impact on Model Performance:}
We next investigate whether the new dataset can
help current models achieve better classification performance.
As a first test, we
use the standard AlexNet \cite{AlexNet} trained on PLACES and compare the model fine-tuned
by the PLACES data \cite{zhou2014learning} with the same model fine-tuned
by both PLACES and LSUN, in both cases fine-tuning on only the 10
categories in LSUN.
As a simple measure to balance the dataset,
we only take at most 200 thousand images from each LSUN category.
% \footnote{We didn't ultilize all data that may enable better
% results, due to limited computation resource we have.}. 
Then we test the classification results on
this 10 categories in PLACES testing set. The error percentage comparison is shown in Figure
~\ref{fig:comp_places_error}. The overall detection error percentage
is 28.6\% with only PLACES and 22.2\% with both PLACES and LSUN, 
which yields a 22.37\% improvement on testing error. Although the
testing is unfavorable to LSUN due to potential dataset bias issues
\cite{datasetBias}, we
can see that the additional of more data can improve the classification
accuracy on PLACES testing set for 8 of the 10 categories.

We then use the pre-trained model to fine-tune on
PASCAL VOC 2012 classification training images with Hinge loss and
evaluate the learned representation. We use two pre-trained models for
comparison. One is AlexNet trained by our object data from
scratch. The other is VGG trained by our object data but initialized
by ImageNet. We sample 300 thousand images
from each of the 20 object categories and use them as training data
from LSUN. To get classification score for
testing images, we sample 10
crops from each testing image at center and four corners with mirroring
and take the average. The results are evaluated on the validation
set for each ablation study. As a comparison, we go through the
same procedure with model pre-trained on ImageNet. The results are
shown in Table~\ref{tab:pascal_val}. The table shows that the model
pre-trained on our data performs significantly better. It indicates
that it is better to have more images in relevant categories than more
categories.

Together with the results on scene categories, it is
interesting to observe that although AlexNet is small compared to
other image classification models developed in recent years, even it
can benefit from more training data.

\newcommand\ver[1]{\rotatebox[origin=c]{90}{#1}}
\begin{table}[hbtp]
%\vspace{-2mm}
\begingroup
\setlength{\tabcolsep}{1.5pt}
\scriptsize
\begin{center}
\begin{tabular}{c|c||c|c|c|c|c|c|c|c|c|c|c|c|c|c|c|c|c|c|c|c||c}
Model & Pre-train  & \ver{aero} & \ver{bike} & \ver{bird} & \ver{boat} & \ver{bottle} &
   \ver{bus} & \ver{car} & \ver{cat} & \ver{chair} & \ver{cow} &
     \ver{table} & \ver{dog} & \ver{horse} & \ver{mbike}
   & \ver{person} &
     \ver{plant} & \ver{sheep} & \ver{sofa} & \ver{train} & \ver{tv} &
       \ver{\,mAP} \\ \hline
AlexNet & ImageNet & 0.96 & 0.86 & 0.88 & 0.85 & 0.49 & 0.94 & 0.75 & 0.89 & 0.62 & 0.86 &
   0.65 & 0.87 & 0.92 & 0.9 & 0.85 & 0.53 & 0.91 & 0.68 & 0.92 &
   0.77 & 0.8 \\ 
AlexNet & LSUN &  0.98 & 0.93 & 0.94 & 0.9 & 0.64 & 0.95 & 0.78 & 0.97 & 0.74 & 0.96
   & 0.72 & 0.96 & 0.98 & 0.94 & 0.85 & 0.59 & 0.96 & 0.76 & 0.97 &
   0.82 & 0.87 \\ 
VGG & ImageNet &   0.95 & 0.85 & 0.92 & 0.86 & 0.65 & 0.92 & 0.8 & 0.93 & 0.68 & 0.77
   & 0.66 & 0.9 & 0.85 & 0.86 & 0.94 & 0.56 & 0.89 & 0.6 & 0.94 &
   0.85 & 0.82 \\ 
VGG & LSUN & 0.98 & 0.92 & 0.94 & 0.9 & 0.69 & 0.96 & 0.82 & 0.96 & 0.75 & 0.97
   & 0.78 & 0.93 & 0.97 & 0.92 & 0.95 & 0.67 & 0.96 & 0.73 & 0.97 &
   0.84 & 0.88 \\ \hline
\end{tabular}
\end{center}
\endgroup
\caption{Comparison of pre-training on ImageNet and LSUN. We pre-train
   AlexNet and VGGnet with ImageNet and our data to compare the
   learned representation. We fine tune the networks after removing
   the last layer on PASCAL VOC 2012 images and compare the
   results. The table shows that pre-training with more images in
   related categories is better than that with more categories.}
\label{tab:pascal_val}
\end{table}

%\vspace{-2mm}
%\subsection{Learned Image Representation}
%\vspace*{-0.05in}

\begin{figure}[hbtp]
\centering
\includegraphics[width=\textwidth]{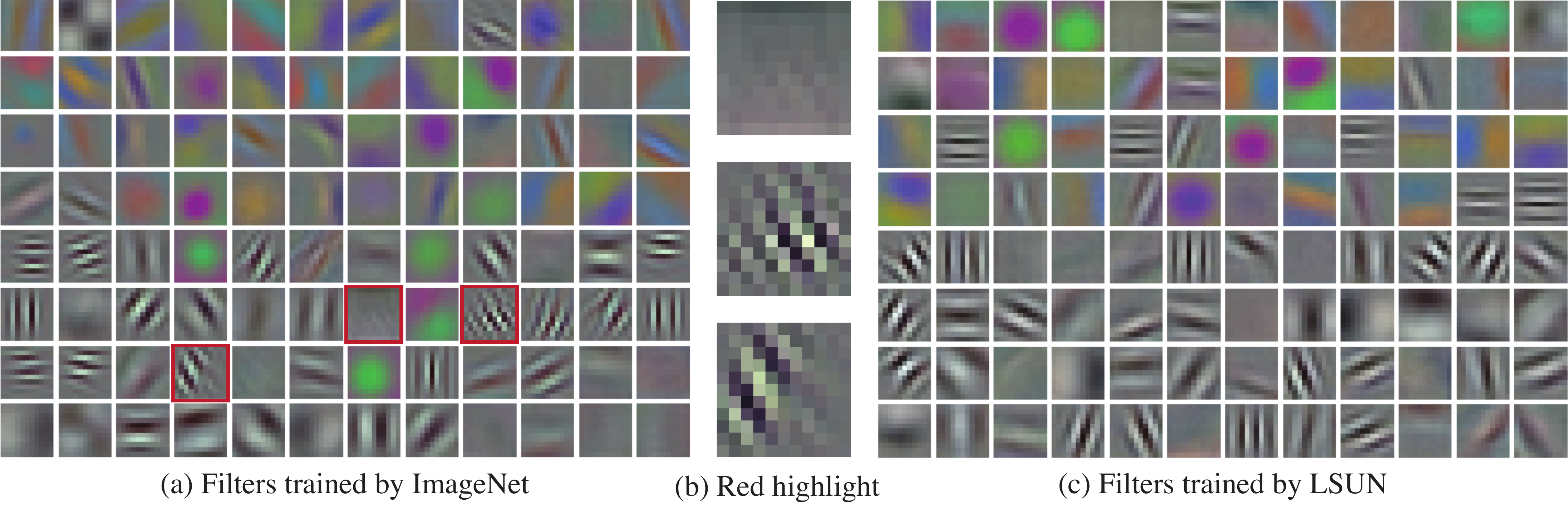}
\caption{Comparison of learned filters. We train AlexNet with our
object images and compare the first level filter with ImageNet. We
observe that the filter pattern learned with our data is cleaner,
while the filters from ImageNet are noisier, as highlighted in (b).}
\label{fig:filter}
\end{figure}

\vspace*{2mm}
\noindent{\bf Learned Image Representation:}
As a final test, we study the image representation learned by our object
categories and compare it to ImageNet to understand the tradeoff
between number of images and categories. We compare AlexNet trained by
ImageNet and our training data used in the last section. Then we check the
response in the first layer. As shown in
Figure~\ref{fig:filter}, we find that the
filters in the first level present cleaner patterns than those learned
on ImageNet data.

\vspace{-0.05in}
\section{Conclusion}
\vspace*{-0.075in}
This paper proposes a working pipeline to acquire large image datasets
with category labels using deep learning with humans in the loop.  We
have constructed an initial version of ``LSUN'' database, a
database with around a million labeled images in each
scene category and more than a million in each of 20 object categories.
Experiments with this dataset have already
demonstrated the great potential of denser training data for visual
recognition.   We will continue to grow
the dataset indefinitely, making it freely available to the community for
further experimentation.

%% \noindent{\bf Acknowledgment.} This work is partially funded by Intel,
%% Google, MERL, Princeton
%% Project X, and a hardware donation from NVIDIA.

% \subsubsection*{Acknowledgments}

% \subsubsection*{References}
{ 
\small
% \footnotesize
  \bibliographystyle{ieee}
  \bibliography{lsun}
}

\end{document}